\documentclass[preprint,12pt,authoryear]{elsarticle}

\usepackage{amssymb}

\usepackage{lineno,hyperref}
\modulolinenumbers[5]
\usepackage{amsmath,amssymb,amsfonts}
\usepackage{algorithmic}
\usepackage{multirow}
\usepackage{graphicx}
\usepackage{textcomp}
\usepackage{floatrow}
\usepackage[center]{caption}
\usepackage{hyperref}
\usepackage{xcolor}
\usepackage{ulem}

\biboptions{authoryear}

\begin{document}

\begin{frontmatter}

\title{Counterfactual Explanation and Instance-Generation using Cycle-Consistent Generative Adversarial Networks}

\author[comsats,ncai]{Tehseen Zia\corref{correspondingauthor}}

\author[comsats,ncai,icube]{Zeeshan Nisar\corref{contribution}}
\author[comsats,ncai,ets]{Shakeeb Murtaza\corref{contribution}}

\affiliation[comsats]{organization={COMSATS University Islamabad, Pakistan}}
\affiliation[ncai]{organization={Medical Imaging and Diagnostics Lab, National Center of Artificial Intelligence, Pakistan}}
\affiliation[icube]{organization={ICube, University of Strasbourg, CNRS (UMR 7357), France}}
\affiliation[ets]{organization={Laboratoire d’imagerie, de vision et d’intelligence artificielle, Ecole de technologie superieure, Montreal, Canada}}

\cortext[correspondingauthor]{Corresponding Author : tehseen.zia@comsats.edu.pk}
\cortext[contribution]{Denotes equal contribution. Ordering determined by random shuffle.}

\begin{abstract}
The image-based diagnosis is now a vital aspect of modern automation assisted diagnosis. To enable models to produce pixel-level diagnosis, pixel-level ground-truth labels are essentially required. However, since it is often not straight forward to obtain the labels in many application domains such as in medical image, classification-based approaches have become the de facto standard to perform the diagnosis. Though they can identify class-salient regions, they may not be useful for diagnosis where capturing all of the evidences is important requirement. Alternatively, a counterfactual explanation (CX) aims at providing explanations using a casual reasoning process of form "If X has not happend, Y would not heppend". Existing CX approaches, however, use classifier to explain features that can change its predictions. Thus, they can only explain class-salient features, rather than entire object of interest. This hence motivates us to propose a novel CX strategy that is not reliant on image classification. This work is inspired from the recent developments in generative adversarial networks (GANs) based image-to-image domain translation, and leverages to translate an abnormal image to counterpart normal image (i.e. counterfactual instance CI) to find discrepancy maps between the two. Since it is generally not possible to obtain abnormal and normal image pairs, we leverage Cycle-Consistency principle (a.k.a CycleGAN) to perform the translation in unsupervised way. We formulate CX in terms of a discrepancy map that, when added from the abnormal image, will make it indistinguishable from the CI. We evaluate our method on three datasets including a synthetic, tuberculosis and BraTS dataset. All these experiments confirm the supremacy of propose method in generating accurate CX and CI.

\end{abstract}


\end{frontmatter}


\section{Introduction}
\label{sec:introduction}
Deep neural network have achieved remarkable competences in many computer vision tasks \cite{1,2,3,4,5}. However,  networks have increasingly become complex and opaque, making it hard to explain and interpret decisions. While various studies have recently addressed the issue of network interpretability, the majority of this activity is focused on explaining a classifier’s decision \cite{6,7,8,9,10}; e.g. by highlighting areas of input image that contribute the most towards the decision. Such studies thus do not consider changes to the input which could lead to the production of different outcomes – i.e. they are neither discriminative nor counterfactual. Casual reasoning beyond correlation is hence crucial for full interpretative explanation of decisions \cite{11,12,13,14}.

\begin{figure}[htbp]
	\begin{center}
		\includegraphics[width=0.6\linewidth]{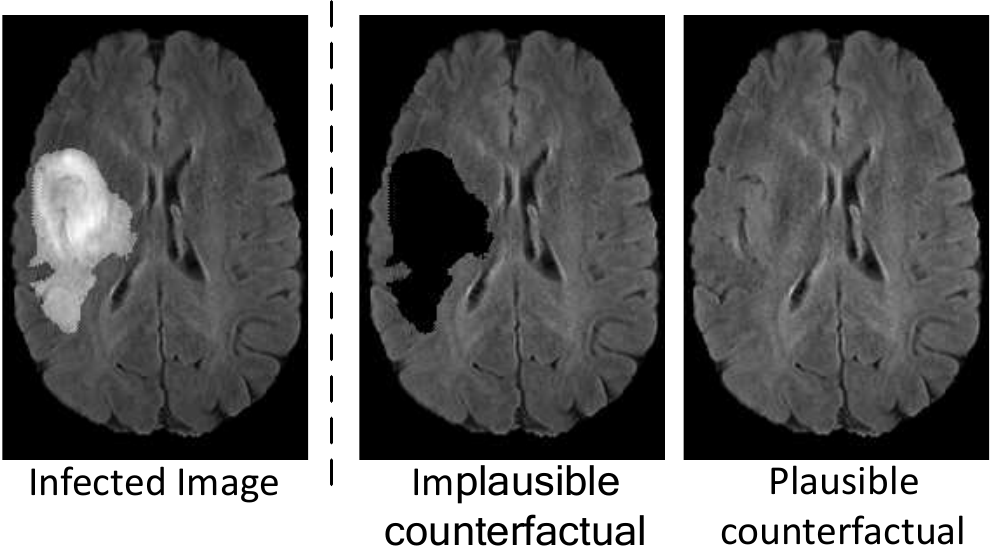}
	\end{center}
	\caption{An illustration of Implausible vs. Plausible counterfactual instances of disease-infected image.}
	\label{fig:Implausible_vs_Plausible}
\end{figure}

A counterfactual explanation (CX) explicates a casual reasoning process of the form: “if X had not happened, Y would not have happened” \cite{11}, for example, “if I hadn’t had these symptoms, I would not have this disease”. Existing de facto state-of art CX methods tend to describe the smallest changes to the input features, producing a counterfactual instance (CI), that alter the prediction of a classifier \cite{13, 14}. However, these CX techniques often produce unrealistic, out-of-distribution CI \ref{fig:Implausible_vs_Plausible} on which the classifiers were not trained. Because classifiers are often easily fooled by unusual input patterns \cite{zee14alcorn2019strike, zee15nguyen2015deep, zee16agarwal2019improving}, we hypothesize that such CI might yield unreliable visual attribution \cite{zee17adebayo2018sanity}. Accompanying CX with a plausible CI thus provide self-explanatory analogy-based explanations. For example, in medical diagnostics, it is useful to address the question “why is this particular disease diagnosed?” by providing an analogy between the input (i.e. “how a scan looks”) and a relevant CI (i.e. “what it should look like”).

The need for this work arises from medical diagnosis where machine diagnosis typically differs from that of human experts in key respects; for instance, a radiologist is trained via observation of many abnormal/normal images such that they are able to transfer their internally-learned representation of the disease to novel image settings. Their training hence enables them to analyze a image by finding abnormalities that differ from a conjectured representation of its CI (i.e. counterpart normal image of the patient). A supervised classification system, by contrast, will typically seek to identify key features indicative of the distinction between normal and abnormal tissue \cite{9}. Inspired by this conjectured expert modus operandi, we seek a methodology capable of the production of a counterfactual image in relation to an input image such that we may use this ‘normal’ image to analyze the input image.

Previous CI-based CX techniques replace a part of an input image (e.g. a square tile) with a specific region of a counterfactual image (i.e. a CI) \cite{12}. Such techniques thus intervene in the original data space, but make only a restricted number of changes. Consequently, the generated CXs are not plausible. Furthermore, these techniques require a dataset consisting in pairs of images from the input and counterfactual classes, which are not always possible to acquire. Here, counterfactual image refers to the normal image against the abnormal input image that help in understanding the cause of disease and also useful for explaining the reason to the patient. Recently, generative models have been used to learn transformations that, when applied to an input image, produce counterfactual images \cite{15}. Since this is one-to-many mapping, specific counterfactual images must be generated randomly from the input image over the transformation set. However, this under-constrained mapping can lead to the production of irrelevant CXs, reflecting undesirable discrepancies between the images. 

This work is primarily inspired from the recent developments in generative adversarial networks (GANs) based medical image translations across modalities and classes \cite{zee18sun2020adversarial, zee19armanious2020medgan, zee20armanious2019unsupervised}. In particular, it is recently demonstrated that GANs \cite{zee21radford2015unsupervised} learn to perform the abnormal-to-normal medical image mapping in unsupervised way via the application of the Cycle-Consistecy GAN \cite{15} principle (a.k.a. CycleGAN). The CycleGAN uses an inverse mapping and cycle consistency (i.e. forwards-backwards) loss to the GAN in order to tackle tasks for which paired training data does not exist, such as abnormal-to-normal medical image pairs. By utilizing and extending this capacity of CycleGAN to produce abnormal-to-normal translation of medical image pairs, we shall demonstrate that it is possible to reformulate CX in terms of a discrepancy map that, when added from the abnormal image, will make it indistinguishable from the counterpart normal image (i.e. CI). To this end, we propose a class of generative models for learning discrepancy maps as a function of abnormal images. In particular, we develop a cascaded model wherein CI of abnormal images are initially generated in the primary phase, such that a generative model is then learned capable of producing a discrepancy map of abnormal w.r.t CI in the second phase. In its final form, we propose a CX-GAN architecture, dubbed Counterfactual Explanation GAN, that learns to generate discrepancy maps simultaneously to learning to perform abnormal-to-normal/CI translation.

Our approach thus aims to improve on a related method proposed in \cite{15}, in which a map is learned that, when added into an abnormal image, renders it indistinguishable from images of the normal class. Since the map-generating function in the \cite{15} case does not aim to produce the CI of an abnormal image but rather any normal-looking image, the learned image translation may depict discrepancies irrelevant to medical diagnosis. We shall, in contrast, set out to constrain the unconstrained abnormal-to-normal image translation function of \cite{15} by generating CI of abnormal images in order to reduce false-positive in CX.

We organize the paper with related work on visual explanation approaches described in Section \ref{sec:relatedwork}, following with the proposed method presented in Section \ref{sec:method}, and experiments/results are reported in Section \ref{sec:experiments}. The paper is concluded in Section \ref{sec:conclusion}.

\section{Related Work}
\label{sec:relatedwork}

Recently, to address the opacity of deep learning, various visual explanation methods have been proposed. Many existing methods are built on convolution neural networks (CNN) and use back-propagation to illuminate class-specific areas of an image \cite{6,7,8,9,10}. A prominent example is the class activation map (CAM) method \cite{9}. CAM uses the global average pooling technique to aggregate feature maps and a fully convolution network (FCN) to localize regions that a network attends to in order to classify the image. Subsequently, the CAM method is improved by replacing global average pooling with a gradient-based feature combination technique. This method is referred as gradient CAM (or gradCAM) \cite{10}. we empirically compare the CAM and gradCAM methods with our proposed method in results section. However, a disadvantage of CAM-based methods is that visual explanation is restricted to the resolution of final layer. Consequently, post-processing is often required to enhance the resolution. A similar style of approach produces visual explanations (in the form of saliency maps) by following gradient signals back to the input image domain. Examples involve Guided Backprop \cite{17}, Excitation Backprop \cite{18} and Integrated Gradients \cite{19}. Several other methods have also been dedicated to explain the decision behind a neural network and to decode its black-box mystery. Layer-wise Relevance Propagation (LRP) \cite{zee1bach2015pixel} is one of these methods that leverage the graph structure of Deep Neural Network (DNN) for a quick and reliable explanation. The authors proposed to deconstruct a non-linear decision by pixel-wise decomposition. For each input feature, LRP is subjected to explain which feature contribute to what extent to a positive or negative decision. LRP operates by propagating the learned prediction function f(x) backwards in the respective DNN, using a set of defined rules. The propagation procedure followed by LRP is subjected to a conservation property, where the decisions are projected down towards the input features. In simple words, what has been received by a neuron in the upper layers must be distributed back to the lower layers in equal amount. This LRP method has been widely used in numerous applications e.g., in explaining a therapy decision \cite{zee2yang2018explaining}, in highlighting the cell structures in microscopy images \cite{zee3binder2018towards}, and in finding the EEG patterns that explain the decision in brain-computer interfaces \cite{zee4sturm2016interpretable}. 

More relevant to our method are counterfactual-based techniques that set out to alter the prediction of the classifier by replacing regions (or pixels) of input images with uninformative values \cite{19,20,21,22}. In \cite{zee12schwab2019cxplain}, a CX approach (a.k.a CXPlain) is proposed to estimate feature importance for machine learning models. The target feature importance is estimated as a decrease in error by adding that feature to the available feature set. The Kullback-Leibler (KL) divergence between target feature importance and predictive feature importance score is then minimized to train CXPlain to predict feature importance. However, in high-dimensional data such as medical images, the model may not be very effective since removing single pixels in high-dimensional images is unlikely to strongly affect a predictive model’s output. As the approach removes features from samples to assess their importance, they often produce unrealistic, out-of-distribution CI on which the classifiers were not trained, which may lead to produce unreliable visual attribution. In \cite{zee13agarwal2020explaining}, the drawback of existing CX approaches to produce unrealistic/ implausible CI is addressed. Instead of removing features (via adding noise, blurring and graying out), an image inpainting based feature removal is proposed to produce plausible CI. However, both \cite{zee12schwab2019cxplain, zee13agarwal2020explaining} approaches rely on a classifier to assess the importance of a removed feature, they are limited to detect certain salient regions rather than entire objects of interest. Recently another approach \cite{zee11cohen2021gifsplanation} has also been proposed to generate counterfactual explanations by increasing or decreasing those features in an input image that causes the prediction. Given any classifier, the authors proposed a simple autoencoder and gradient update (Latent Shift) that may exaggerate or limit the features utilized for prediction by transforming the latent representation of a given input image. Though this approach addresses the limitation of generative models that are monolithic, difficult and time consuming to train because of their hyper-parameter settings and loss function stability, this approach is limited  by the latent representation of the autoencoder. If the decoder isn't descriptive enough, it won't be able to appropriately interpret the classifier's features. An iterative procedure is adopted to replace the input regions in \cite{14}. A generative model is employed to fill-in masked image from unmasked images in \cite{14}. In \cite{13}, minimal regions of an input are obtained that maximally alter a classifier’s score both positively and negatively.

Unlike the above counterfactual techniques, we do not set out to describe those regions of an image that contribute most to the prediction of the classifier. Rather, we will aim to explain an input image with respect to a counterfactual image. Previous work to this end includes a patch-based editing model \cite{12}  trained to replace regions of an input image with counterfactual images that alter the decision of the classifier. However, the method requires a paired dataset of input/output images, which may not always be possible to acquire. In \cite{15}, a generative model is used to learn changes to an input image (i.e. CX) that make it indistinguishable from any random counterfactual image. However, due to an unconstrained map function, the method also generates undesirable effects (i.e. false positive CXs) when attempting to describe irrelevant discrepancies between the input and a random counterfactual image. The examples of these undesirable effects are depicted in the experimental section.  

\begin{figure*}[htbp]
	\begin{center}
		\includegraphics[width=1\linewidth,trim=7 7 7 7, clip]{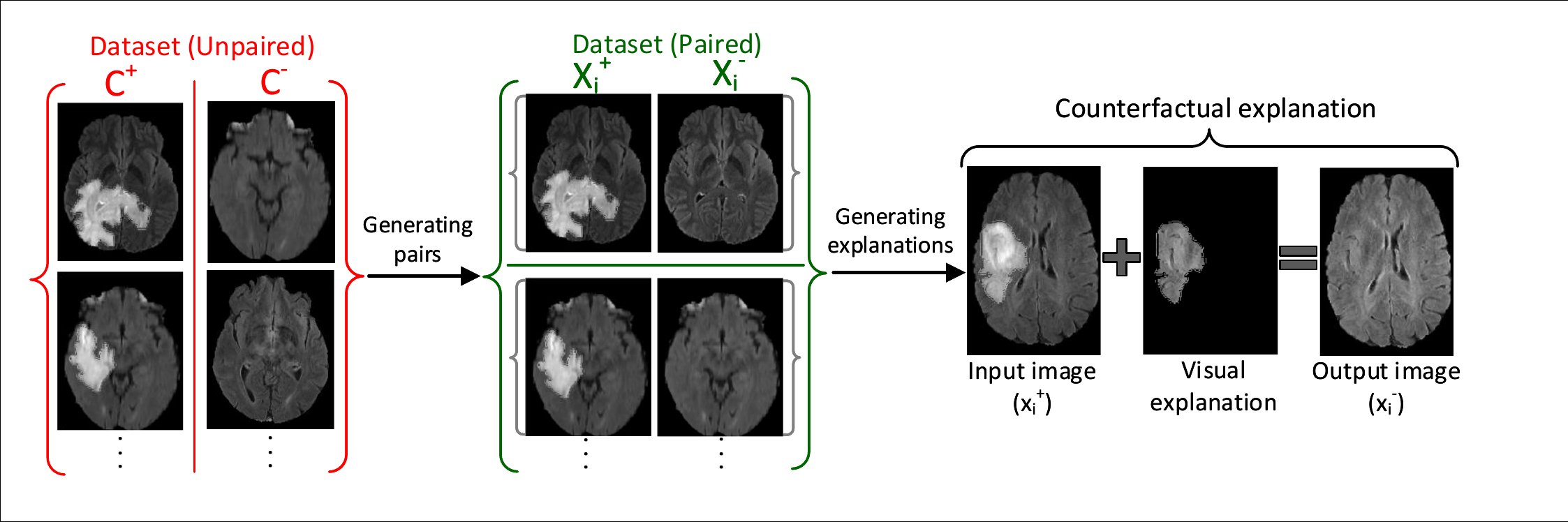}
	\end{center}
	\caption{Schematic diagram of cascaded CX-GAN model.}
	\label{fig:schemati_diagram_cascaded}
\end{figure*}

\section{Methodology}
\label{sec:method}
Consider a two class problem $c\in{0,1}$, a positive $c^+$ and a negative class $c^-$, and instance of positive class $x^+$ and negative class $x^-$ are drawn from distributions $p_d(c^+)$ and $p_d(c^-)$ respectively. We assume that for any instance $x^+in$ class $c^+$, there exist a counterpart instance $x^-$ in class $c^-$ such that $x^+$ differs from $x^-$ merely based on class-specific features. Further, the counterpart instances of positive and negative classes are not accessible. Such a problem setting is readily available in medical imaging domains where positive and negative classes represent unhealthy and healthy scans, respectively. Given this setting, we seek to achieve two objectives: 1) to build a model architecture to generate counterpart negative instance from positive instance. We refer to counterpart negative instance of a positive instance as counterfactual instance CI. 2) To build a model architecture to explain an instance of positive class with reference to its CI. We refer to this explanation as counterfactual explanation CX. To generate CI, we need a domain adaptation model that can translate a positive instance into CI. We use CycleGAN for this purpose due to its state-of-the-art performance in unsupervised domain adaptation, especially to translate medical images across domains and classes. On generating CI, one can simply formulate the CX as a difference between positive instance and CI. However, this would require a post-processing step to define an effective threshold function. To circumvent post-processing, we formulate CX problem as follows: given an instance of positive class $x^+$, generates a CX (i.e., visual attribution map) $M(x+)$, that when added to $x^+$ produces $x^-$:

\begin{equation} \label{zeeeq1}
    x^- = x^+ + M(x^+)
\end{equation}

Ideally, $x^+$ merely differs from $x^-$  in class specific features, thereby $M(x^+)$ contains all the features which distinguish $x^+$ from its counterfactual instance ${x}^-$. For example, in medical images, M will by definition contain the effects of a disease visible in the images, i.e. a disease effect map. In the following sections, we describe two approaches to learn models to generate CI and CX.

\begin{figure*}[htbp]
	\begin{center}
		\includegraphics[width=1\linewidth,trim=7 7 7 7, clip]{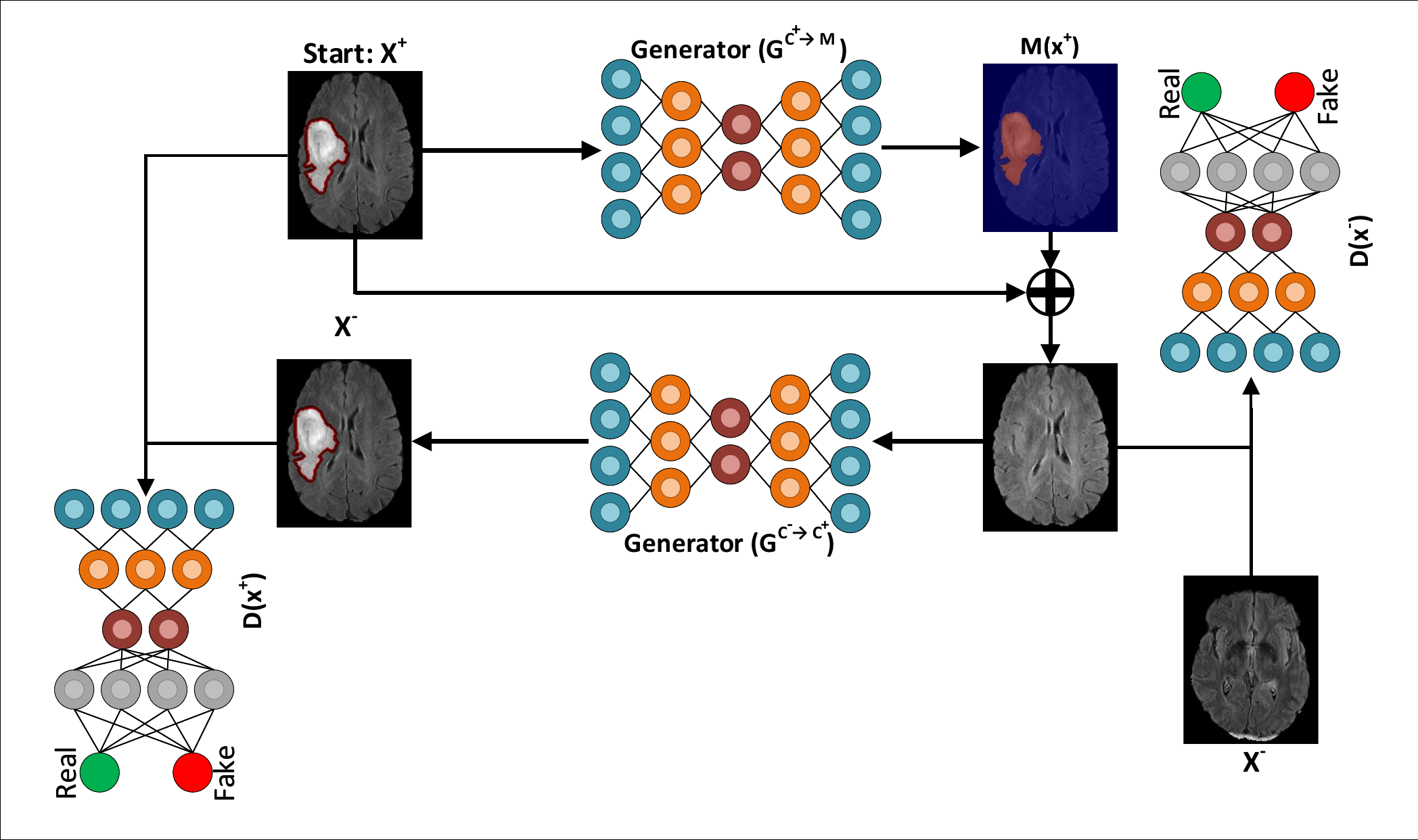}
	\end{center}
	\caption{Block diagram of the integrated CX-GAN consists of two pairs of generator and discriminator, sharing a similar architecture. The generator is a convolution network with skip connections between encoder and decoder layers. Similarly, the discriminator is also a convolution classifier and neurons at the output layer.}
	\label{fig:block_diagram_integrated}
\end{figure*}

\subsection{Cascaded Approach}\label{sec:problem_for_cacaded}
This approach is a two-phase procedure, hence we refer to it as “cascaded” approach. In the first phase, we learn to generate CI of positive instance. We employ CycleGAN to perform this unsupervised positive-to-counterfactual image translation due to its superior performance in similar tasks. In the second phase, we construct a dataset of positive and counterfactual pairs $(x^+, x^-)$ to learn $M(x^+)$ using Equation \ref{zeeeq1}. Although, various model are proposed for supervised domain translation, we have adopted a well-established model pix2pix GAN. 

\noindent
\textbf{Counterfactual Instance Generation: }
As described above, we use CycleGAN to generate CI of positive class instance. The main component of the model is a generator $\mathcal{G}^{c^+\rightarrow c^-}$, that takes as input an instance of positive class, and produces as output CI. To converge the learning process, the generator must produce realistic CI $x^-={\mathcal{G}^{c^+\rightarrow c^-}(x}^+)$, capable of fooling negative discriminator ${\mathcal{D}}^{c^-}$. The cycle consistency regularization is leveraged via another generator $\mathcal{G}^{c^-\rightarrow c^+}$ and discriminator ${\ \mathcal{D}}^{c^+}$ in order to constraint the model to produce counterfactual.  The CI generation can thus be defined as an objective function ${\ \mathcal{L}}_{CI}$, consisting of two parts: a GAN loss ${\mathcal{L}\ }_{GAN}$ and a cyclic-consistency loss ${\mathcal{L}\ }_{CC}$. Mathematically,

\begin{equation}
    {\ \mathcal{L}}_{CI}={\mathcal{L}\ }_{GAN}+{\lambda\mathcal{L}\ }_{CC}
\end{equation}

Where $\mathcal{L}_{GAN}$ is used to simultaneously train both generators $\mathcal{G}^{c^+\rightarrow c^-}$ and $\mathcal{G}^{c^-\rightarrow c^+}$ and is defined as follows:
\begin{multline}\label{eq:10}
    {\mathcal{L}\ }_{GAN}=\mathbb{E}_{c^+}\left[\ln{{\ \mathcal{D}}^{c^+}}\left(x^+\right)\right]+\mathbb{E}_{c^-}\left[\ln{{\ \mathcal{D}}^{c^-}}\left(x^-\right)\right]\\
    +\mathbb{E}_{c^-}\left[\ln{{(1-D}^{c^+}}\left(\mathcal{G}^{c^-\rightarrow c^+}(x^-)\right))\right]+\mathbb{E}_{c^+}\left[\ln{{(1-D}^{c^-}}\left(\mathcal{G}^{c^+\rightarrow c^-}(x^+)\right))\right]
\end{multline}

This is a joint loss function of both forward cycleGAN and backward cycleGAN used in CycleGAN.
and ${\mathcal{L}\ }_{CC}$ is defined as follows:
\begin{multline}\label{eq:11}
    {\mathcal{L}\ }_{CC}=\mathbb{E}_+\left[\left|\left|\mathcal{G}^{c^-\rightarrow c^+}\left(\mathcal{G}^{c^+\rightarrow c^-}\left(x^+\right)\right)-x^+\right|\right|_1\right]+\\ \mathbb{E}_-[\left|\left|\mathcal{G}^{c^+\rightarrow c^-}\left(\mathcal{G}^{c^-\rightarrow c^+}\left(x^-\right)\right)-x^-\right|\right|_1] 
\end{multline}

The first term of ${\mathcal{L}}_{CC}$ is a forward cycle-consistency which aims to bring back $x^+$ to the original form after translating it into counterfactual $x^-$, i.e. $x^+\rightarrow\mathcal{G}^{c^+\rightarrow c^-}\left(x^+\right)\rightarrow\ \mathcal{G}^{c^-\rightarrow c^+}\left(\mathcal{G}^{c^+\rightarrow c^-}\left(x^+\right)\right)$. The second is a backward cycle-consistency which aims to reproduce $x^-$ after translating it into $x^+, i.e. x^-\rightarrow\mathcal{G}^{c^-\rightarrow c^+}\left(x^-\right)\rightarrow\ \mathcal{G}^{c^+\rightarrow c^-}\left(\mathcal{G}^{c^-\rightarrow c^+}\left(x^-\right)\right)$.  The optimization of ${\mathcal{L}\ }_{CC}$ ensures that a positive instance is translated into counterfactual, rather than any instance of negative class. 

\noindent
\textbf{Counterfactual Explanation Generation}
In this phase, we want to generate CX of positive instance with respect to its CI. Such an explanation is vital in medical images to visualize features of unhealthy scan with reference to the counterpart healthy scan.  To achieve the goal, a simple way is to reformulate Equation 1 as ${M\left(x^+\right)=x}^- - x^+$, and pose CX as a difference between positive instance and its CI. However, the method would require a post-processing step to define an effective threshold function. To cope with disadvantage, we propose a residual GAN (RGAN) to explain class-specific features of positive instance. We adopt pix2pix GAN to build the model \cite{27}; however, in contrast to the standard pix2pix GAN, we use a generator $\mathcal{G}^{c^+\rightarrow M}$ to take an input image $x^+$ and produce a map $M\left(x^+\right)$ that when added into $x^+$ generates a counterfactual $x^-$ (in practice $x^-)$. The discriminator ${\mathcal{D}}^{c^-}$ is used to discriminate between real $x^-$ and generated $x^-$. To train the RGAN, we define residual L1 loss $\mathcal{L}_{1r}$ and residual adversarial loss functions $\mathcal{L}_{rGAN}$ by adopting L1 and LGAN loss functions proposed in \cite{27}.

\begin{equation}
    \mathcal{L}_{1r}=\mathbb{E}_{c^+,c^-}{||(x^--\left(x^++M\left(x^+\right)\right)||}_1
\end{equation}

\begin{equation}
    \mathcal{L}_{rGAN}=\ \mathbb{E}_{c^-}\left[\ln{{\ \mathcal{D}}^{c^-}}\left(x^-\right)+\ln{(1-{\ \mathcal{D}}^{c^-}}({x^++\mathcal{G}}^{c^+\rightarrow M}\left(x^+\right)))\right]
\end{equation}

We adopt second term of the losses to train forward cycleGAN to produce CX (i.e. visual attribution) map. Finally we optimize RGAN with the following min-max objective function:

\begin{equation}
    G^\ast=arg{min}_G{min}_D\mathcal{L}_{rGAN}+\lambda\mathcal{L}_{1r}
\end{equation}

A schematic diagram of the method is shown in Figure \ref{fig:schemati_diagram_cascaded}. 

\subsection{End-to-End Integrated Approach}
\label{sec:problem_for_integrated_model}
A disadvantage of the cascaded model is that separate networks are trained to generate CI and CX, and the performance of CX network relies on efficacy of the CI network. This section presents a method for joint learning of both CI and CX through an integrated model. We build on our CI model and enables it to learn a map $M(x^+)$ to transform a positive instance $x^+$ into a counterfactual $x^-$ using Equation 1 through the mapping $x^++M\left(x^+\right)\rightarrow\ x^-$. To achieve that, we replace generator $\mathcal{G}^{c^+\rightarrow c^-}$ with a generator $\mathcal{G}^{c^+\rightarrow M}$ to learn the function $M(x^+)$ in the forward consistency cycle. It is important to note that we learn $M(x^+)$ using CycleGAN, rather than pix2pix GAN as in Section \ref{sec:problem_for_cacaded}. This is mainly because we do not have positive and counterfactual pairs, unlike in Section \ref{sec:problem_for_cacaded}  where we construct such a dataset using counterfactual instance generation model.  To train this integrated end-to-end CI and CX model, we modify the GAN loss function ${\mathcal{L}\ }_{GAN}$ and cyclic-consistency loss ${\mathcal{L}\ }_{CC}$ given in Equation \ref{eq:10} and \ref{eq:11} to define CX GAN (CX-GAN) loss function ${\mathcal{L}\ }_{cxGAN}$ and CX cyclic-consistency loss ${\mathcal{L}}_{cxCC}$ respectively as follows:

\begin{multline}\label{zeeeq2}
    {\mathcal{L}\ }_{cxGAN}=\mathbb{E}_{c^+}\left[\ln{D^{c^+}}\left(x^+\right)\right]+\mathbb{E}_{c^-}\left[\ln{{\ \mathcal{D}}^{c^-}}\left(x^-\right)\right]+\\ \mathbb{E}_{c^-}\left[\ln{{(1-D}^{c^+}}\left(\mathcal{G}^{c^-\rightarrow c^+}(x^-)\right))\right]+\mathbb{E}_{c^+}\left[\ln{{(1-D}^{c^-}}\left(x^++\mathcal{G}^{c^+\rightarrow M}(x^+)\right))\right]
\end{multline}

\begin{multline}\label{zeeeq3}
    {\mathcal{L}\ }_{cxCC}=\mathbb{E}{c^+}+\left[\left|\left|\mathcal{G}^{c^-\rightarrow c^+}\left(x^++\mathcal{G}^{c^+\rightarrow M}\left(x^+\right)\right)-x^+\right|\right|_1\right]+ \\  \mathbb{E}_{c^-}[\left|\left|x^++\mathcal{G}^{c^+\rightarrow M}\left(\mathcal{G}^{c^-\rightarrow c^+}\left(x^-\right)\right)-x^-\right|\right|_1]
\end{multline}

Note that Equation \ref{zeeeq2} and \ref{zeeeq3} differs from Equation \ref{eq:10} and \ref{eq:11} respectively in last term, to enable forward cycle to generate CX map. Finally we define CX loss ${\mathcal{L}\ }_{cx}$ to optimize CX-GAN as follows:

\begin{equation}
    {\mathcal{L}\ }_{cx}=\ {\mathcal{L}\ }_{cxGAN}+\lambda\mathcal{L}_{cxCC}
\end{equation}

Once we train CX-GAN, we only keep the generator $\mathcal{G}^{c^+\rightarrow M}$ and throwaway generator $\mathcal{G}^{c^-\rightarrow c^+}$ and discriminators $D^{c^+}$ and $D^{c^-}$. We input an instance of positive class to the network $\mathcal{G}^{c^+\rightarrow M}$ and obtain a CX map $M(x^+)$.

\subsection{Network Training}
\label{sec:network_training}
The training a of GANs aims to find a Nash equilibrium in two-players non-cooperative game. This is unfortunately a very hard problem with no reliable solution. Consequently, GANs’ training may end up in a failure mode rather than convergence. This incurs training instability in GANs. We used a recommended techniques from previous work to ensure stable GAN training; we reduce model oscillation \cite{5} by updating discriminators using a history of generated images rather than the ones produced by the latest generators, as suggested in \cite{16, shrivastava2017learning}. We use 50 generated images to update the discriminator as suggested in \cite{15}.

All the networks are optimized with the ADAM-optimizer with momentum parameters: $\beta_1=0.5$, $\beta_2=0.9$. The learning rate and batch size are kept as 0.0002 and 1, respectively. The stopping criterion is chosen to be a patience of 10 epochs of validating the precision of generating a normal image pair. 
We train networks on a GPU-based desktop system with 128 GB RAM, Nvidia TitanX Pascal (12 GB VRAM) and 10 core Intel Xeon processor. 

\section{Experiments}\label{sec:experiments}
We perform experiments on a synthetic dataset and two publically available medical imaging datasets including BraTS and tuberculosis datasets (i.e. Shenzhen, Montgomery County and Korean Institute of Tuberculosis). We evaluate our proposed method against comparable visual explanation methods including CAM \cite{9}, gradCAM \cite{10}, and VA-GAN \cite{15}, where CAM and gradCAM use classification networks, while VA-GAN and the proposed CX-GAN employ image translation networks. For simplicity, all networks are built with a similar architecture to that used for the discriminator of the proposed method. However, for CAM methods, we replace the last two layers with a global average pooling layer followed by a dense prediction to create class-specific activation maps for a visual explanation, as described in \cite{9}. We use Dice-Coefficient, Intersection over Union (IoU) and normalized cross correlation (NCC) as our evaluation metrics.

Implementation details and code of proposed methodology is available at \underline
{\url{https://github.com/zeeshannisar/CX_GAN}}

\subsection{Experiments on Synthetic Data}
\subsubsection{Synthetic dataset \cite{15}}
\begin{figure}[htbp]
	\begin{center}
		\includegraphics[width=0.6\linewidth]{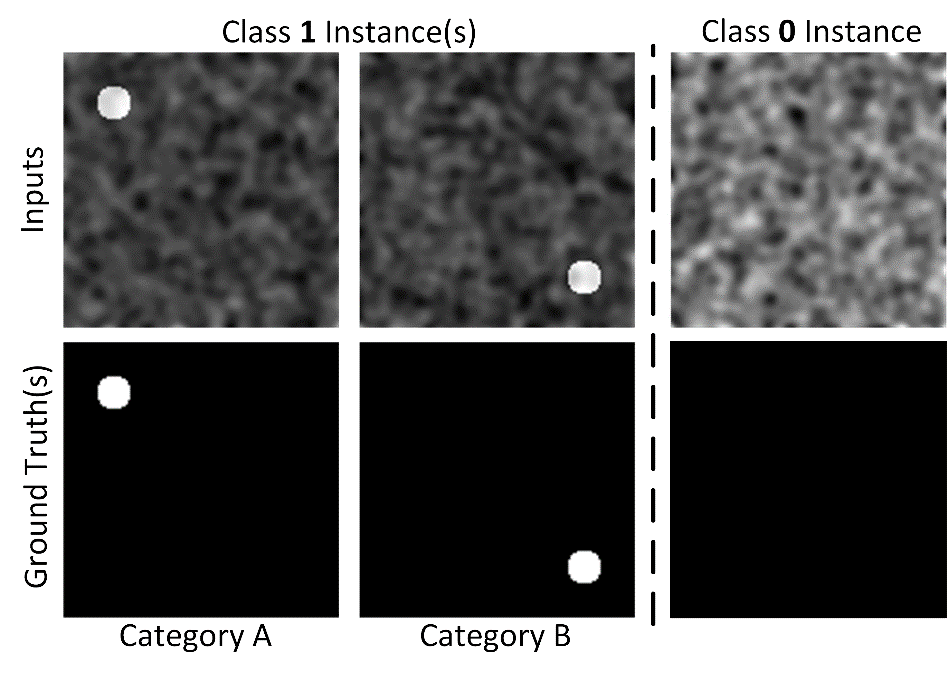}
	\end{center}
	\caption{Examples of synthetic data. Left of the dotted line are samples of Class 1 (i.e. the disease class) and right of the dotted line are samples of Class 0 (i.e. the normal class). The upper row shows the input and the bottom row shows the ground truth.}
	\label{fig:example_synthetic_data}
\end{figure}

In addition to real medical imaging datasets, we evaluate both the proposed and related methods on a synthetically generated dataset of 10000 128x128 images classified into two classes. One half of the dataset represents a healthy control group (label 0) and another half represents a patient group (label 1). The dataset is generated by close adherence to the data generation process set out in \cite{15}. Images of the healthy control group are generated by convolving a Gaussian blurring filter with random iid Gaussian noise. Images of the patient control group are also produced from noise, but they also contain effects attributable to one of two diseases. These effects are introduced by inserting a circle on the top left side of the image (disease A), or a circle at the bottom right side (disease B). Note that both diseases share the same label. The circles are randomly placed with a  maximum 5-pixel offset in each direction (this effect is  introduced to making the problem more difficult; however it has no effect on the outcome). Samples images are shown in Figure \ref{fig:example_synthetic_data}.

\subsubsection{Evaluation}
We divide the overall data set into 80-20 train and test set following \cite{15}. For quantitative evaluation, we calculate IoU and Dice score between the disease maps and the visual explanation. We use the maximum pixel value as a threshold to covert the visual explanation map into binary mask. Following \cite{15}, we also employed the normalized cross correlation (NCC) measure between the ground-truth maps and the predicted visual explanation maps.

\subsubsection{Results}
\begin{figure}[htbp]
	\begin{center}
		\includegraphics[width=1\linewidth]{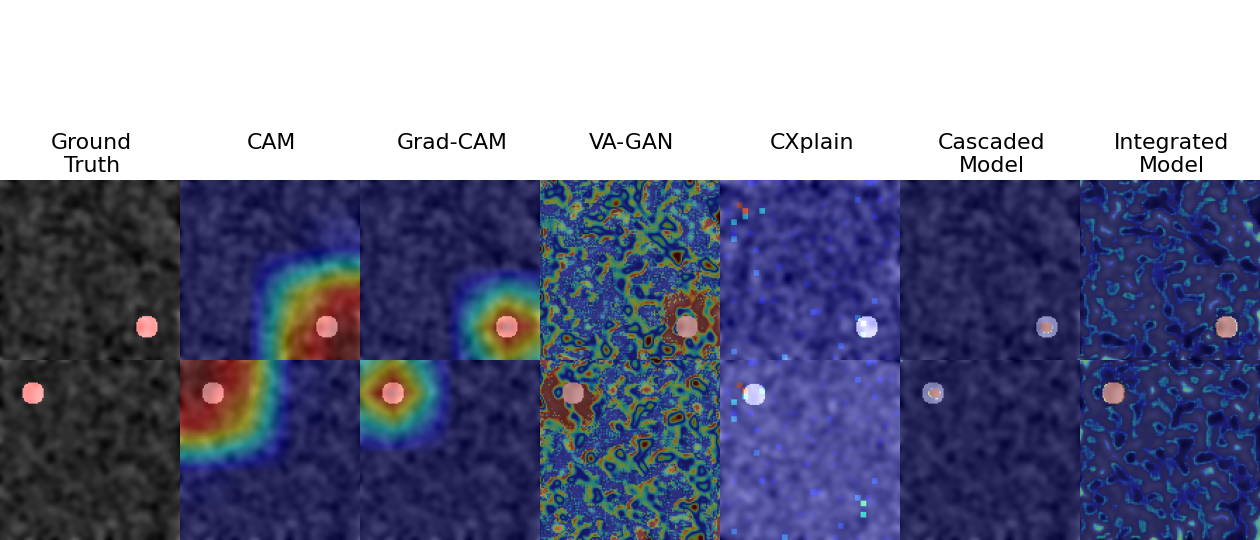}
	\end{center}
	\caption{Examples of visualization maps of compared methods on synthetic data.}
	\label{fig:result_synthetic_data}
\end{figure}

Quantitative results for all of the methods on the synthetic data are reported in Table \ref{tab:scores_syntheticdata}. The results clearly indicate the relative supremacy of the proposed method; examples of visual explanation maps for all of the methods are shown in Figure \ref{fig:result_synthetic_data}. It is apparent that the CAM-based methods tend to focus on areas where the circles are distributed uniformly, and are unable to provide fine-grained visualization maps (the effect can clearly be observed from the visualization map of the CAM-based methods in Figure \ref{fig:result_synthetic_data}).

It is further apparent that VA-GAN produces noisy visualization maps due to its under-constrained mapping from unaligned noisy images; the noisy maps contain many false positives which degrade VA-GAN performance. (this effect can be seen in the visual explanation map of VA-GAN from Figure \ref{fig:result_synthetic_data}). Contrarily, the proposed method produces far more plausible visual explanation maps primarily due to the constrained CycleGAN-based mapping, CX-GAN can thus better describe the input image w.r.t. the generated CI. Finally, we notice that the integrated CX-GAN shows a slight improvement over cascaded CX-GAN. We shall hence report the performance of integrated CX-GANs in subsequent experiments. 

\begin{table}[htbp]
	\centering
	\begin{tabular}{|p{3.3cm}|c|c|c|c|}
		\hline
		\multicolumn{1}{|c|}{\multirow{2}{*}{Method}} & \multirow{2}{*}{IoU Score} & \multirow{2}{*}{Dice Score} & \multicolumn{2}{c|}{NCC Score} \\ \cline{4-5} 
		\multicolumn{1}{|c|}{} &  &  & Mean & Std \\ \hline
		CAM & 10.4 & 18.8 & 0.29 & 0.0025 \\ \hline
		GradCAM & 30.7 & 47 & 0.6 & 0.0482 \\ \hline
		VA-GAN & 87.2 & 92.8 & 0.92 & 0.0091 \\ \hline
		CXplain & 12.32 & 27.88 & -- & -- \\ \hline
		Cascaded CX-GAN & 90. 10 & 93.6 & 0.91 & 0.009 \\ \hline
		Integrated CX-GAN & 91.4 & 95.5 & 0.95 & 0.0061 \\ \hline
	\end{tabular}
	\caption{IoU, Dice Scores and NCC Scores of evaluated methods on synthetic data.}
	\label{tab:scores_syntheticdata}
\end{table}

\subsection{Experiments on Medical Imaging Data}
\subsubsection{Datasets }
\noindent
\textbf{Tuberculosis chest X-ray dataset \cite{zee5kim2016deconvolutional}}:
These datasets contain de-identified CXRs of normal and TB-effected cases. The study is conducted on three diverse public resources: (1) the National Institute of Health (NIH) Tuberculosis Chest X-ray database \cite{zee7jaeger2014two}, (2) the Belarus Tuberculosis database \cite{zee9b}, and (3) Korean Institute of Tuberculosis (KIT) under Korean National Tuberculosis Association, South Korea \cite{zee8ryoo2014activities}. The NIH is further categorized into two separate datasets: (a) Montgomery County (MC) and (b) Shenzhen. The Montgomery and Shenzhen dataset contains 138 and 662 patients respectively, with and without TB. The MC Dataset consists of 138 CXRs including 80 normal and 58 TB-effected CXRs. The Shenzhen dataset comprises 662 CXRs where 326 are normal, and 336 are TB-infected. The Belarus dataset has a total of 304 CXRs of TB-infected patients. The KIT dataset contains 10, 848 DICOM images with 7,020 normal and 3,828 TB-infected CXRs. We follow the experimental protocol of \cite{zee5kim2016deconvolutional}. The input data is preprocessed with the following steps: (1) a border from the edges of each CXR is cropped to exempt pixels with a high noise-ratio, (2) each CXR is resized to $527\times 527$ for computational efficiency. We crop 15 pixels randomly to retain the morphology of TB markers. Any additional augmentations (except for horizontal mirroring and flipping) allowable for lesion deformation are not adopted. In the final step, we normalize CXRs with z-score normalization. 

\noindent
\textbf{BraTS datasets \cite{zee6bakas2017advancing}}:The dataset contains brain MRIs classified into normal and tumorous classes. We preprocess the data to filter-out MRI slices that contain the full brain. The dataset contains 3174 images where 2711 are tumorous and 463 non-tumorous. We split each set into 80-20 train/test sets, resulting in 2538 training images and 636 testing images. The filtered slices are resized to $256\times 256$ and the data normalized to the 0-to-1 range. We further increase the data size by performing run-time augmentation on training sets through random jittering and mirroring. For augmenting, the images are scaled to $286\times 286$ and then randomly cropped to $256\times 256$.
\begin{table}[!b]
	\centering
	\resizebox{0.9\textwidth}{!}{%
		\begin{tabular}{|l|c|c|c|c|}
			\hline
			& \multicolumn{2}{c|}{\textbf{Tuberculosis dataset}} & \multicolumn{2}{c|}{\textbf{BraTS dataset}} \\ \hline
			\multicolumn{1}{|c|}{\textbf{Method}} & \textbf{IoU Score} & \textbf{Dice Score} & \textbf{IoU Score} & \textbf{Dice Score} \\ \hline
			CAM & 19.67 & 28.56 & 30.8 & 45.1 \\ \hline
			gradCAM & 32.33 & 45.19 & 54.7 & 60.3 \\ \hline
			VA-GAN & 29.15 & 53.91 & 89.5 & 93.2 \\ \hline
			CXplain & -- & -- & 14.21 & 24.74 \\ \hline
			Cascaded CX-GAN & 61.4 & 73.3 & 77.1 & 81.2 \\ \hline
			Integrated CX-GAN & 74.73 & 81.47 & 91.4 & 94 \\ \hline
		\end{tabular}%
	}
	\caption{IoU and Dice Scores of evaluated methods on tuberculosis and BraTS datasets}
	\label{tab:score_brats_tb}
\end{table}

\subsubsection{Evaluation}
We use the visual explanation maps generated by the networks for semantic segmentation of disease affected regions. We split datasets into 80-20 train/test sets. To gauge the efficacy of the respective networks, we employ mean IoU and Dice coefficient (i.e. the standard metrics to evaluate semantic segmentation methods). To calculate these metrics, we convert visual explanation maps into binary masks. The highest value in the explanation map is used as a threshold to convert the visual explanation map into binary mask. For the tuberculosis dataset, we obtain \textit{pixel-level} ground-truths of the Shenzhen dataset provided by the authors of \cite{4}. In the tuberculosis evaluation, the test set is based on the Shenzhen as pixel-level ground-truths are only available for this data-set.   

\subsubsection{Results}
\begin{figure}[htbp]
	\begin{center}
		\includegraphics[width=0.9\linewidth]{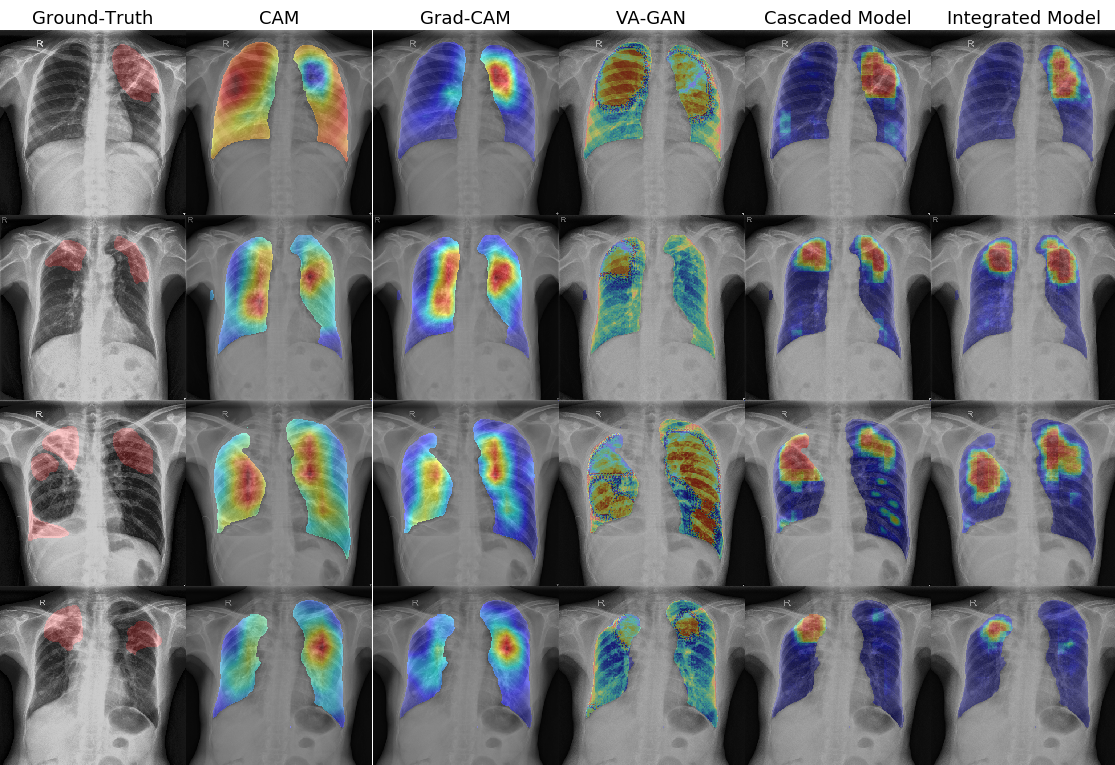}
	\end{center}
	\caption{Examples of visualization maps of compared methods on tuberculosis data.}
	\label{fig:result_tb}
\end{figure}

\begin{figure}[htbp]
	\begin{center}
		\includegraphics[width=0.9\linewidth]{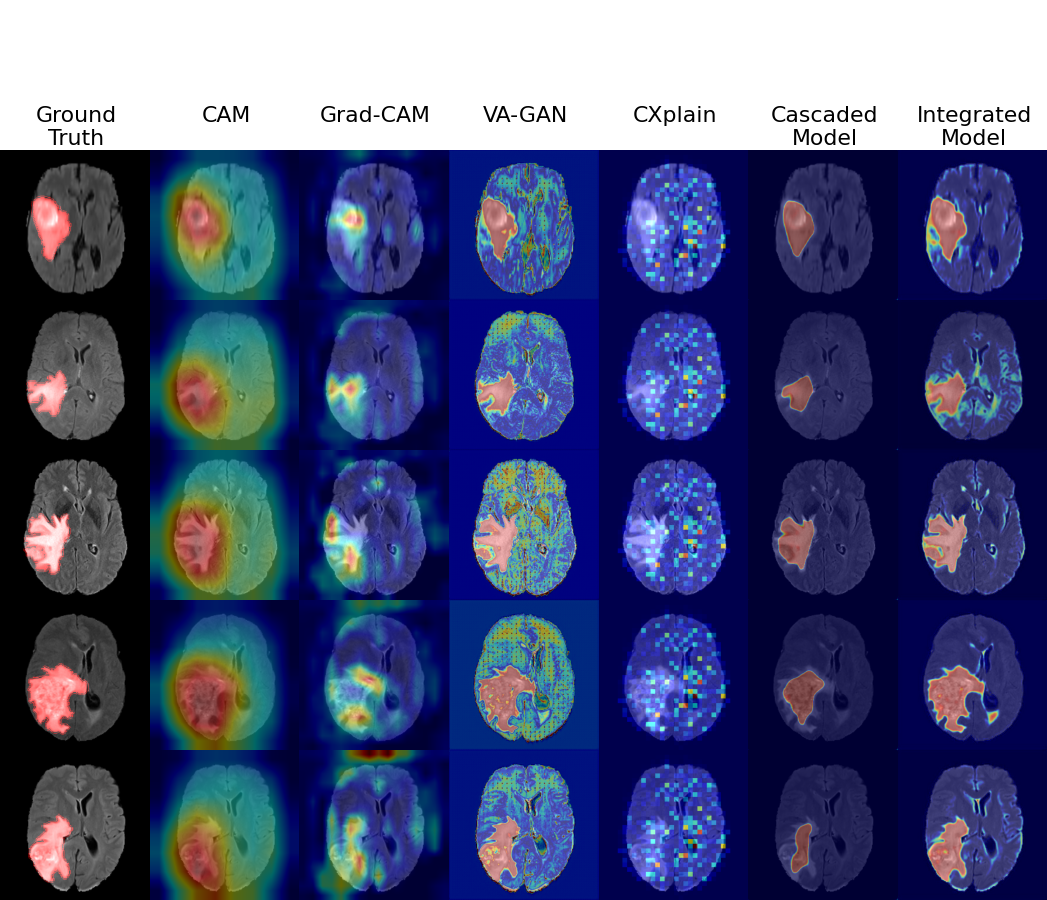}
	\end{center}
	\caption{Examples of visualization maps of the compared methods on BraTS data.}
	\label{fig:result_brats}
\end{figure}

Table \ref{tab:score_brats_tb} shows quantitative results of these experiments. The proposed method significantly outperforms the other methods. The methods in general perform better on the BraTS data than the tuberculosis data because BraTS contains easily identifiable tumorous regions as compared with barely-recognizable tuberculosis markers. Examples of the visual explanation maps on the tuberculosis and BraTS data are depicted in Figure \ref{fig:result_tb} and \ref{fig:result_brats} respectively. 

The results and findings are consistent with the synthetic data. We believe that CAM-based methods show limited performance due to focusing only on a minimal set of the most discriminative features while disregarding the rest. The visual explanation of CAM method is hence noisy, low resolution and often disoriented. gradCAM improves the explanation of CAM method in terms of noise and resolution. However, the gradCAM explanation region is much smaller than ground truth. Similarly to the synthetic data, the VA-GAN explanation map is quite noisy and difficult to comprehend. As mentioned above, this likely happens because VA-GAN is trained to map a set of images from one domain to a random permutation of images in another domain. Due to the under-constrained mapping, VA-GAN hence takes into account undesirable features (i.e. false positives) introducing unnecessary discrepancies between input and output images. Contrarily, it can be observed from the figures that the visual explanation map of proposed method is neither smaller in magnitude nor noisier than the ground truth, and is hence more plausible compared to other methods.

\subsection{Evaluation of Generated Counterfactual Instances with Non-Resemblance Measure}
We measure the quality of generated CI on BraTS dataset by measuring the non-resemblance between input image (i.e. tumorous image) and generated CI (i.e. normal image). We compute non-resemblance between images separately for corresponding tumorous and normal regions. The tumorous regions are separated using the ground truth (GT). The non-resemblance score is calculated as follows:

\begin{multline}
Tumerous\ regions\ =\ \ 1-\left(\frac{1}{N_i}\ \sum_{i}^{n_i}{({y_i-\ x}_i)}\ \right)\  \\ \therefore i:n_i\ index\ where\ GT\ ==1
\end{multline}

\begin{multline}
Normal\ regions\ =\ \ 1-\left(\frac{1}{N_j}\ \sum_{j}^{n_j}{({y_j-\ x}_j)}\ \right)\ \ \\ \therefore j:n_j\ index\ where\ GT\ ==0
\end{multline}

\begin{figure}[htbp]
	\begin{center}
		\includegraphics[width=0.7\linewidth]{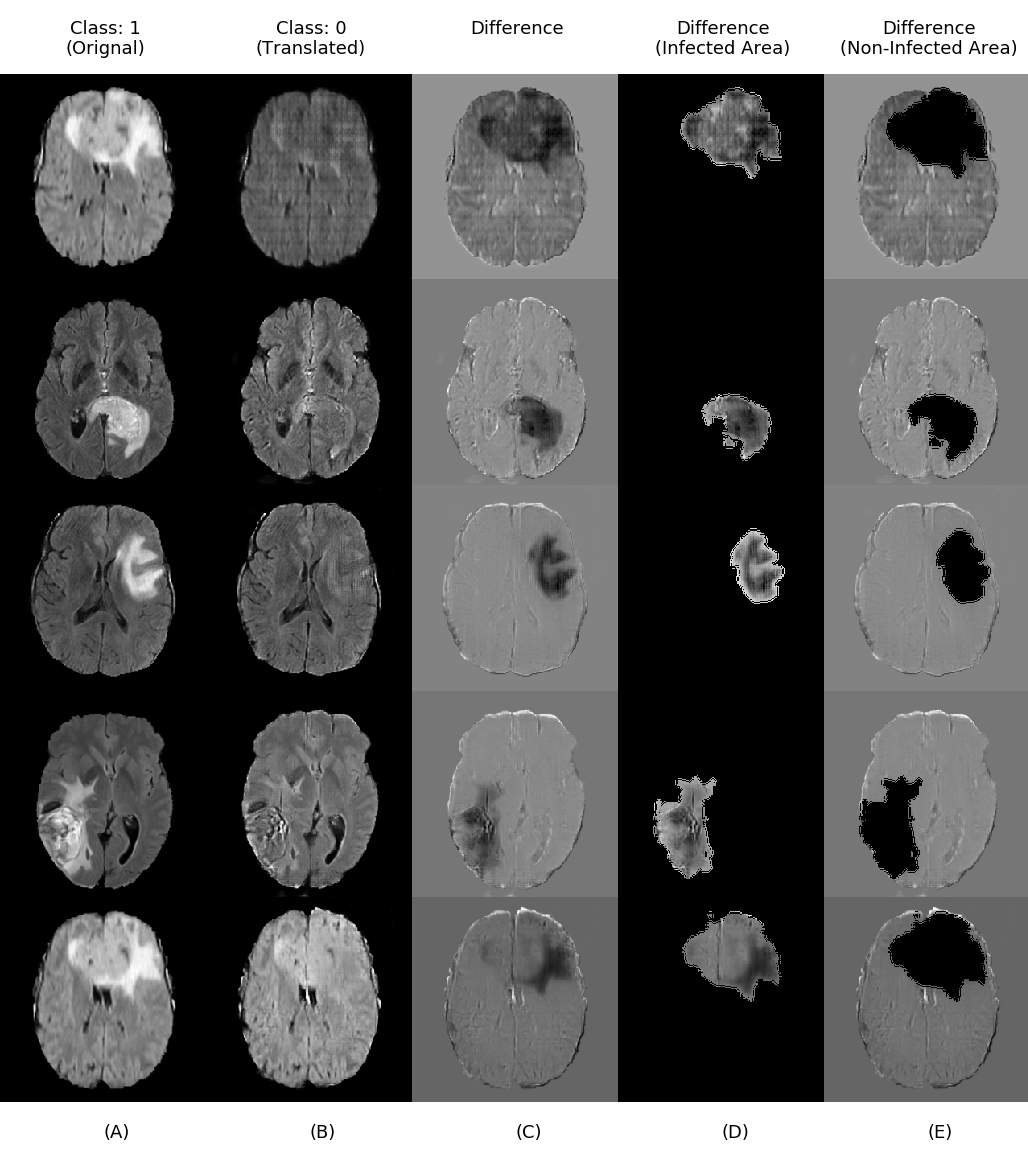}
	\end{center}
	\caption{Illustration of the non-resemblance score measuring process.  A) input image (i.e. tumorous image), B) generated CI (i.e. normal image), C) difference (or non-resemblance) image, D) separated tumorous region and E) separated normal region.}
	\label{fig:result_non_resumlance}
\end{figure}

The process of computing the non-resemblance score is depicted in Figure \ref{fig:result_non_resumlance}. The quantitative results are shown in Table \ref{tab:Non_Resemblance_Score}.  The proposed CX-GAN shows a low non-resemblance (i.e. high similarity) score between normal regions of the images; that is, the model does not change these regions while generating the CI. On the other hand, the non-resemblance between tumorous regions is quite high (i.e. low similarity) since the model has eliminated tumorous markers while generating the CI (i.e. normal image). In contrast, VA-GAN shows a high non-resemblance score between normal regions of the images and a low non-resemblance between tumorous regions.

\begin{table}[htbp]
	\centering
	\resizebox{0.8\textwidth}{!}{%
		\begin{tabular}{|l|c|c|c|}
			\hline
			\multicolumn{1}{|c|}{\multirow{2}{*}{\textbf{Method}}} & \multicolumn{3}{c|}{\textbf{Non-Resemblance Score}} \\ \cline{2-4} 
			\multicolumn{1}{|c|}{} & \multicolumn{1}{l|}{\textbf{Tumorous region}} & \multicolumn{1}{l|}{\textbf{Normal region}} & \multicolumn{1}{l|}{\textbf{Total Non-Resemblance}} \\ \hline
			CX-GAN & 0.67 & 0.33 & 0.5 \\ \hline
			VA-GAN & 0.44 & 0.68 & 0.56 \\ \hline
		\end{tabular}%
	}
	\caption{Non-Resemblance Score on Brats data for generated pairs}
	\label{tab:Non_Resemblance_Score}
\end{table}

\subsection{Evaluation of Generated Counterfactual Instances with Structural Similarity Index Measure}
\begin{figure}[htbp]
	\begin{center}
		\includegraphics[width=0.6\linewidth]{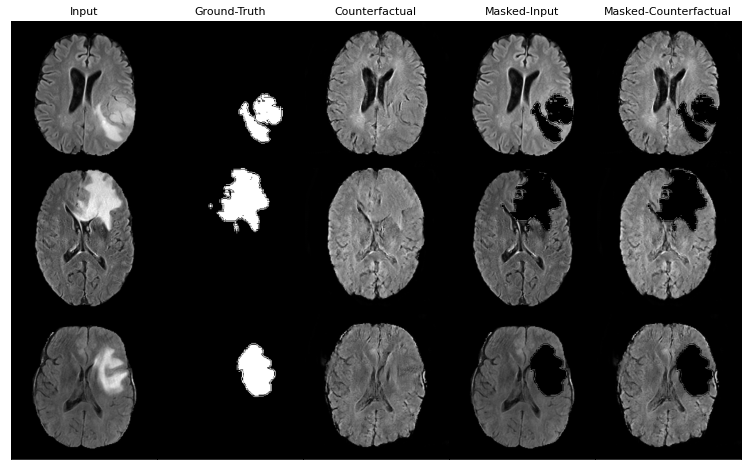}
	\end{center}
	\caption{Illustration of the structural similarity index score measuring process.  A) input image (i.e. tumorous image), B) ground-truth, C) generated CI (i.e. normal image), D) masked-input (with removed diseased region), D) masked generated CI (with removed diseased region).}
	\label{fig:result_ssim}
\end{figure}

To quantify the effectiveness of our proposed approach we did not only assessed the physical consistency with segmentation masks using IoU and Dice-Score, we also assessed the structural similarity in terms of preserving the input image structure with in the generated CI. We compute the structural similarity index measure (SSIM) \cite{zee10wang2004image} between the masked-input image (i.e. image with removed diseased region) and masked generated CI (i.e. normal image with removed diseased region) as shown in Figure \ref{fig:result_ssim}. 

The diseased bio-markers are removed from both of the images based on the available ground-truths. The quantitative results are reported in Table \ref{tab:SSIM_Score} and with these scores we can clearly  assume that our approach is enforcing to preserve the input image structural similarity with its generated CI. 

\begin{table}[htbp]
	\centering
	\resizebox{0.4\textwidth}{!}
	{
		\begin{tabular}{|c|c|c|}
			\hline
			 & BRaTs Data & Synthetic Data \\ \hline
	    SSIM & 0.7459 & 0.6701 \\ \hline
		\end{tabular}
	}
	\caption{SSIM (close to 1 means perfect match) between masked-input and its generated masked CI on Brats data and Synthetic data}
	\label{tab:SSIM_Score}
\end{table}

\section{Conclusion}\label{sec:conclusion}
A counterfactual explanation (CX) and counterfactual instance (CI) generation method is developed; we first we outlined a cascaded model for individual learning of CX and CI generation. Then, we developed an integrated model for joint learning of both CXs and CIs. The models are built on cycle-consistent GANs. We show that, by contrast with back-propagation-based and pre-existing counterfactual-based techniques, the proposed method produces significantly more refined CXs and plausible CIs.

\bibliography{cas-refs}

\begin{thebibliography}{45}
\expandafter\ifx\csname natexlab\endcsname\relax\def\natexlab#1{#1}\fi
\providecommand{\url}[1]{\texttt{#1}}
\providecommand{\href}[2]{#2}
\providecommand{\path}[1]{#1}
\providecommand{\DOIprefix}{doi:}
\providecommand{\ArXivprefix}{arXiv:}
\providecommand{\URLprefix}{URL: }
\providecommand{\Pubmedprefix}{pmid:}
\providecommand{\doi}[1]{\href{http://dx.doi.org/#1}{\path{#1}}}
\providecommand{\Pubmed}[1]{\href{pmid:#1}{\path{#1}}}
\providecommand{\bibinfo}[2]{#2}
\ifx\xfnm\relax \def\xfnm[#1]{\unskip,\space#1}\fi
\bibitem[{Krizhevsky et~al.(2012)Krizhevsky, Sutskever, and Hinton}]{1}
\bibinfo{author}{A.~Krizhevsky}, \bibinfo{author}{I.~Sutskever},
  \bibinfo{author}{G.~E. Hinton},
\newblock \bibinfo{title}{Imagenet classification with deep convolutional
  neural networks},
\newblock in: \bibinfo{booktitle}{Advances in neural information processing
  systems}, \bibinfo{year}{2012}, pp. \bibinfo{pages}{1097--1105}.
\bibitem[{Ronneberger et~al.(2015)Ronneberger, Fischer, and Brox}]{2}
\bibinfo{author}{O.~Ronneberger}, \bibinfo{author}{P.~Fischer},
  \bibinfo{author}{T.~Brox},
\newblock \bibinfo{title}{U-net: Convolutional networks for biomedical image
  segmentation},
\newblock in: \bibinfo{booktitle}{International Conference on Medical image
  computing and computer-assisted intervention},
  \bibinfo{organization}{Springer}, \bibinfo{year}{2015}, pp.
  \bibinfo{pages}{234--241}.
\bibitem[{Long et~al.(2015)Long, Shelhamer, and Darrell}]{3}
\bibinfo{author}{J.~Long}, \bibinfo{author}{E.~Shelhamer},
  \bibinfo{author}{T.~Darrell},
\newblock \bibinfo{title}{Fully convolutional networks for semantic
  segmentation},
\newblock in: \bibinfo{booktitle}{Proceedings of the IEEE conference on
  computer vision and pattern recognition}, \bibinfo{year}{2015}, pp.
  \bibinfo{pages}{3431--3440}.
\bibitem[{Ren et~al.(2015)Ren, He, Girshick, and Sun}]{4}
\bibinfo{author}{S.~Ren}, \bibinfo{author}{K.~He},
  \bibinfo{author}{R.~Girshick}, \bibinfo{author}{J.~Sun},
\newblock \bibinfo{title}{Faster r-cnn: Towards real-time object detection with
  region proposal networks},
\newblock in: \bibinfo{booktitle}{Advances in neural information processing
  systems}, \bibinfo{year}{2015}, pp. \bibinfo{pages}{91--99}.
\bibitem[{Goodfellow et~al.(2014)Goodfellow, Pouget-Abadie, Mirza, Xu,
  Warde-Farley, Ozair, Courville, and Bengio}]{5}
\bibinfo{author}{I.~Goodfellow}, \bibinfo{author}{J.~Pouget-Abadie},
  \bibinfo{author}{M.~Mirza}, \bibinfo{author}{B.~Xu},
  \bibinfo{author}{D.~Warde-Farley}, \bibinfo{author}{S.~Ozair},
  \bibinfo{author}{A.~Courville}, \bibinfo{author}{Y.~Bengio},
\newblock \bibinfo{title}{Generative adversarial networks},
\newblock in: \bibinfo{booktitle}{Advances in neural information processing
  systems}, \bibinfo{year}{2014}, pp. \bibinfo{pages}{2672--2680}.
\bibitem[{Simonyan et~al.(2013)Simonyan, Vedaldi, and Zisserman}]{6}
\bibinfo{author}{K.~Simonyan}, \bibinfo{author}{A.~Vedaldi},
  \bibinfo{author}{A.~Zisserman},
\newblock \bibinfo{title}{Deep inside convolutional networks: Visualising image
  classification models and saliency maps},
\newblock \bibinfo{journal}{arXiv preprint arXiv:1312.6034}
  (\bibinfo{year}{2013}).
\bibitem[{Zeiler and Fergus(2014)}]{7}
\bibinfo{author}{M.~D. Zeiler}, \bibinfo{author}{R.~Fergus},
\newblock \bibinfo{title}{Visualizing and understanding convolutional
  networks},
\newblock in: \bibinfo{booktitle}{European conference on computer vision},
  \bibinfo{organization}{Springer}, \bibinfo{year}{2014}, pp.
  \bibinfo{pages}{818--833}.
\bibitem[{Yosinski et~al.(2015)Yosinski, Clune, Nguyen, Fuchs, and Lipson}]{8}
\bibinfo{author}{J.~Yosinski}, \bibinfo{author}{J.~Clune},
  \bibinfo{author}{A.~Nguyen}, \bibinfo{author}{T.~Fuchs},
  \bibinfo{author}{H.~Lipson},
\newblock \bibinfo{title}{Understanding neural networks through deep
  visualization},
\newblock \bibinfo{journal}{arXiv preprint arXiv:1506.06579}
  (\bibinfo{year}{2015}).
\bibitem[{Zhou et~al.(2016)Zhou, Khosla, Lapedriza, Oliva, and Torralba}]{9}
\bibinfo{author}{B.~Zhou}, \bibinfo{author}{A.~Khosla},
  \bibinfo{author}{A.~Lapedriza}, \bibinfo{author}{A.~Oliva},
  \bibinfo{author}{A.~Torralba},
\newblock \bibinfo{title}{Learning deep features for discriminative
  localization},
\newblock in: \bibinfo{booktitle}{Proceedings of the IEEE conference on
  computer vision and pattern recognition}, \bibinfo{year}{2016}, pp.
  \bibinfo{pages}{2921--2929}.
\bibitem[{Selvaraju et~al.(2017)Selvaraju, Cogswell, Das, Vedantam, Parikh, and
  Batra}]{10}
\bibinfo{author}{R.~R. Selvaraju}, \bibinfo{author}{M.~Cogswell},
  \bibinfo{author}{A.~Das}, \bibinfo{author}{R.~Vedantam},
  \bibinfo{author}{D.~Parikh}, \bibinfo{author}{D.~Batra},
\newblock \bibinfo{title}{Grad-cam: Visual explanations from deep networks via
  gradient-based localization},
\newblock in: \bibinfo{booktitle}{Proceedings of the IEEE International
  Conference on Computer Vision}, \bibinfo{year}{2017}, pp.
  \bibinfo{pages}{618--626}.
\bibitem[{Molnar(2019)}]{11}
\bibinfo{author}{C.~Molnar}, \bibinfo{title}{Interpretable machine learning},
  \bibinfo{publisher}{Lulu. com}, \bibinfo{year}{2019}.
\bibitem[{Goyal et~al.(2019)Goyal, Wu, Ernst, Batra, Parikh, and Lee}]{12}
\bibinfo{author}{Y.~Goyal}, \bibinfo{author}{Z.~Wu},
  \bibinfo{author}{J.~Ernst}, \bibinfo{author}{D.~Batra},
  \bibinfo{author}{D.~Parikh}, \bibinfo{author}{S.~Lee},
\newblock \bibinfo{title}{Counterfactual visual explanations},
\newblock \bibinfo{journal}{arXiv preprint arXiv:1904.07451}
  (\bibinfo{year}{2019}).
\bibitem[{Dhurandhar et~al.(2018)Dhurandhar, Chen, Luss, Tu, Ting, Shanmugam,
  and Das}]{13}
\bibinfo{author}{A.~Dhurandhar}, \bibinfo{author}{P.-Y. Chen},
  \bibinfo{author}{R.~Luss}, \bibinfo{author}{C.-C. Tu},
  \bibinfo{author}{P.~Ting}, \bibinfo{author}{K.~Shanmugam},
  \bibinfo{author}{P.~Das},
\newblock \bibinfo{title}{Explanations based on the missing: Towards
  contrastive explanations with pertinent negatives},
\newblock in: \bibinfo{booktitle}{Advances in Neural Information Processing
  Systems}, \bibinfo{year}{2018}, pp. \bibinfo{pages}{592--603}.
\bibitem[{Chang et~al.(2018)Chang, Creager, Goldenberg, and Duvenaud}]{14}
\bibinfo{author}{C.-H. Chang}, \bibinfo{author}{E.~Creager},
  \bibinfo{author}{A.~Goldenberg}, \bibinfo{author}{D.~Duvenaud},
\newblock \bibinfo{title}{Explaining image classifiers by counterfactual
  generation},
\newblock \bibinfo{journal}{arXiv preprint arXiv:1807.08024}
  (\bibinfo{year}{2018}).
\bibitem[{Alcorn et~al.(2019)Alcorn, Li, Gong, Wang, Mai, Ku, and
  Nguyen}]{zee14alcorn2019strike}
\bibinfo{author}{M.~A. Alcorn}, \bibinfo{author}{Q.~Li},
  \bibinfo{author}{Z.~Gong}, \bibinfo{author}{C.~Wang},
  \bibinfo{author}{L.~Mai}, \bibinfo{author}{W.-S. Ku},
  \bibinfo{author}{A.~Nguyen},
\newblock \bibinfo{title}{Strike (with) a pose: Neural networks are easily
  fooled by strange poses of familiar objects},
\newblock in: \bibinfo{booktitle}{Proceedings of the IEEE/CVF Conference on
  Computer Vision and Pattern Recognition}, \bibinfo{year}{2019}, pp.
  \bibinfo{pages}{4845--4854}.
\bibitem[{Nguyen et~al.(2015)Nguyen, Yosinski, and Clune}]{zee15nguyen2015deep}
\bibinfo{author}{A.~Nguyen}, \bibinfo{author}{J.~Yosinski},
  \bibinfo{author}{J.~Clune},
\newblock \bibinfo{title}{Deep neural networks are easily fooled: High
  confidence predictions for unrecognizable images},
\newblock in: \bibinfo{booktitle}{Proceedings of the IEEE conference on
  computer vision and pattern recognition}, \bibinfo{year}{2015}, pp.
  \bibinfo{pages}{427--436}.
\bibitem[{Agarwal et~al.(2019)Agarwal, Nguyen, and
  Schonfeld}]{zee16agarwal2019improving}
\bibinfo{author}{C.~Agarwal}, \bibinfo{author}{A.~Nguyen},
  \bibinfo{author}{D.~Schonfeld},
\newblock \bibinfo{title}{Improving robustness to adversarial examples by
  encouraging discriminative features},
\newblock in: \bibinfo{booktitle}{2019 IEEE International Conference on Image
  Processing (ICIP)}, \bibinfo{organization}{IEEE}, \bibinfo{year}{2019}, pp.
  \bibinfo{pages}{3801--3505}.
\bibitem[{Adebayo et~al.(2018)Adebayo, Gilmer, Muelly, Goodfellow, Hardt, and
  Kim}]{zee17adebayo2018sanity}
\bibinfo{author}{J.~Adebayo}, \bibinfo{author}{J.~Gilmer},
  \bibinfo{author}{M.~Muelly}, \bibinfo{author}{I.~Goodfellow},
  \bibinfo{author}{M.~Hardt}, \bibinfo{author}{B.~Kim},
\newblock \bibinfo{title}{Sanity checks for saliency maps},
\newblock \bibinfo{journal}{arXiv preprint arXiv:1810.03292}
  (\bibinfo{year}{2018}).
\bibitem[{Baumgartner et~al.(2018)Baumgartner, Koch, Can~Tezcan, Xi~Ang, and
  Konukoglu}]{15}
\bibinfo{author}{C.~F. Baumgartner}, \bibinfo{author}{L.~M. Koch},
  \bibinfo{author}{K.~Can~Tezcan}, \bibinfo{author}{J.~Xi~Ang},
  \bibinfo{author}{E.~Konukoglu},
\newblock \bibinfo{title}{Visual feature attribution using wasserstein gans},
\newblock in: \bibinfo{booktitle}{Proceedings of the IEEE Conference on
  Computer Vision and Pattern Recognition}, \bibinfo{year}{2018}, pp.
  \bibinfo{pages}{8309--8319}.
\bibitem[{Sun et~al.(2020)Sun, Wang, Huang, Ding, Greenspan, and
  Paisley}]{zee18sun2020adversarial}
\bibinfo{author}{L.~Sun}, \bibinfo{author}{J.~Wang},
  \bibinfo{author}{Y.~Huang}, \bibinfo{author}{X.~Ding},
  \bibinfo{author}{H.~Greenspan}, \bibinfo{author}{J.~Paisley},
\newblock \bibinfo{title}{An adversarial learning approach to medical image
  synthesis for lesion detection},
\newblock \bibinfo{journal}{IEEE journal of biomedical and health informatics}
  \bibinfo{volume}{24} (\bibinfo{year}{2020}) \bibinfo{pages}{2303--2314}.
\bibitem[{Armanious et~al.(2020)Armanious, Jiang, Fischer, K{\"u}stner, Hepp,
  Nikolaou, Gatidis, and Yang}]{zee19armanious2020medgan}
\bibinfo{author}{K.~Armanious}, \bibinfo{author}{C.~Jiang},
  \bibinfo{author}{M.~Fischer}, \bibinfo{author}{T.~K{\"u}stner},
  \bibinfo{author}{T.~Hepp}, \bibinfo{author}{K.~Nikolaou},
  \bibinfo{author}{S.~Gatidis}, \bibinfo{author}{B.~Yang},
\newblock \bibinfo{title}{Medgan: Medical image translation using gans},
\newblock \bibinfo{journal}{Computerized medical imaging and graphics}
  \bibinfo{volume}{79} (\bibinfo{year}{2020}) \bibinfo{pages}{101684}.
\bibitem[{Armanious et~al.(2019)Armanious, Jiang, Abdulatif, K{\"u}stner,
  Gatidis, and Yang}]{zee20armanious2019unsupervised}
\bibinfo{author}{K.~Armanious}, \bibinfo{author}{C.~Jiang},
  \bibinfo{author}{S.~Abdulatif}, \bibinfo{author}{T.~K{\"u}stner},
  \bibinfo{author}{S.~Gatidis}, \bibinfo{author}{B.~Yang},
\newblock \bibinfo{title}{Unsupervised medical image translation using
  cycle-medgan},
\newblock in: \bibinfo{booktitle}{2019 27th European Signal Processing
  Conference (EUSIPCO)}, \bibinfo{organization}{IEEE}, \bibinfo{year}{2019},
  pp. \bibinfo{pages}{1--5}.
\bibitem[{Radford et~al.(2015)Radford, Metz, and
  Chintala}]{zee21radford2015unsupervised}
\bibinfo{author}{A.~Radford}, \bibinfo{author}{L.~Metz},
  \bibinfo{author}{S.~Chintala},
\newblock \bibinfo{title}{Unsupervised representation learning with deep
  convolutional generative adversarial networks},
\newblock \bibinfo{journal}{arXiv preprint arXiv:1511.06434}
  (\bibinfo{year}{2015}).
\bibitem[{Springenberg et~al.(2014)Springenberg, Dosovitskiy, Brox, and
  Riedmiller}]{17}
\bibinfo{author}{J.~T. Springenberg}, \bibinfo{author}{A.~Dosovitskiy},
  \bibinfo{author}{T.~Brox}, \bibinfo{author}{M.~Riedmiller},
\newblock \bibinfo{title}{Striving for simplicity: The all convolutional net},
\newblock \bibinfo{journal}{arXiv preprint arXiv:1412.6806}
  (\bibinfo{year}{2014}).
\bibitem[{Zhang et~al.(2018)Zhang, Bargal, Lin, Brandt, Shen, and
  Sclaroff}]{18}
\bibinfo{author}{J.~Zhang}, \bibinfo{author}{S.~A. Bargal},
  \bibinfo{author}{Z.~Lin}, \bibinfo{author}{J.~Brandt},
  \bibinfo{author}{X.~Shen}, \bibinfo{author}{S.~Sclaroff},
\newblock \bibinfo{title}{Top-down neural attention by excitation backprop},
\newblock \bibinfo{journal}{International Journal of Computer Vision}
  \bibinfo{volume}{126} (\bibinfo{year}{2018}) \bibinfo{pages}{1084--1102}.
\bibitem[{Sundararajan et~al.(2017)Sundararajan, Taly, and Yan}]{19}
\bibinfo{author}{M.~Sundararajan}, \bibinfo{author}{A.~Taly},
  \bibinfo{author}{Q.~Yan},
\newblock \bibinfo{title}{Axiomatic attribution for deep networks},
\newblock in: \bibinfo{booktitle}{Proceedings of the 34th International
  Conference on Machine Learning-Volume 70}, \bibinfo{organization}{JMLR. org},
  \bibinfo{year}{2017}, pp. \bibinfo{pages}{3319--3328}.
\bibitem[{Bach et~al.(2015)Bach, Binder, Montavon, Klauschen, M{\"u}ller, and
  Samek}]{zee1bach2015pixel}
\bibinfo{author}{S.~Bach}, \bibinfo{author}{A.~Binder},
  \bibinfo{author}{G.~Montavon}, \bibinfo{author}{F.~Klauschen},
  \bibinfo{author}{K.-R. M{\"u}ller}, \bibinfo{author}{W.~Samek},
\newblock \bibinfo{title}{On pixel-wise explanations for non-linear classifier
  decisions by layer-wise relevance propagation},
\newblock \bibinfo{journal}{PloS one} \bibinfo{volume}{10}
  (\bibinfo{year}{2015}) \bibinfo{pages}{e0130140}.
\bibitem[{Yang et~al.(2018)Yang, Tresp, Wunderle, and
  Fasching}]{zee2yang2018explaining}
\bibinfo{author}{Y.~Yang}, \bibinfo{author}{V.~Tresp},
  \bibinfo{author}{M.~Wunderle}, \bibinfo{author}{P.~A. Fasching},
\newblock \bibinfo{title}{Explaining therapy predictions with layer-wise
  relevance propagation in neural networks},
\newblock in: \bibinfo{booktitle}{2018 IEEE International Conference on
  Healthcare Informatics (ICHI)}, \bibinfo{organization}{IEEE},
  \bibinfo{year}{2018}, pp. \bibinfo{pages}{152--162}.
\bibitem[{Binder et~al.(2018)Binder, Bockmayr, H{\"a}gele, Wienert, Heim,
  Hellweg, Stenzinger, Parlow, Budczies, Goeppert
  et~al.}]{zee3binder2018towards}
\bibinfo{author}{A.~Binder}, \bibinfo{author}{M.~Bockmayr},
  \bibinfo{author}{M.~H{\"a}gele}, \bibinfo{author}{S.~Wienert},
  \bibinfo{author}{D.~Heim}, \bibinfo{author}{K.~Hellweg},
  \bibinfo{author}{A.~Stenzinger}, \bibinfo{author}{L.~Parlow},
  \bibinfo{author}{J.~Budczies}, \bibinfo{author}{B.~Goeppert}, et~al.,
\newblock \bibinfo{title}{Towards computational fluorescence microscopy:
  Machine learning-based integrated prediction of morphological and molecular
  tumor profiles},
\newblock \bibinfo{journal}{arXiv preprint arXiv:1805.11178}
  (\bibinfo{year}{2018}).
\bibitem[{Sturm et~al.(2016)Sturm, Lapuschkin, Samek, and
  M{\"u}ller}]{zee4sturm2016interpretable}
\bibinfo{author}{I.~Sturm}, \bibinfo{author}{S.~Lapuschkin},
  \bibinfo{author}{W.~Samek}, \bibinfo{author}{K.-R. M{\"u}ller},
\newblock \bibinfo{title}{Interpretable deep neural networks for single-trial
  eeg classification},
\newblock \bibinfo{journal}{Journal of neuroscience methods}
  \bibinfo{volume}{274} (\bibinfo{year}{2016}) \bibinfo{pages}{141--145}.
\bibitem[{Fong and Vedaldi(2017)}]{20}
\bibinfo{author}{R.~C. Fong}, \bibinfo{author}{A.~Vedaldi},
\newblock \bibinfo{title}{Interpretable explanations of black boxes by
  meaningful perturbation},
\newblock in: \bibinfo{booktitle}{Proceedings of the IEEE International
  Conference on Computer Vision}, \bibinfo{year}{2017}, pp.
  \bibinfo{pages}{3429--3437}.
\bibitem[{Dabkowski and Gal(2017)}]{21}
\bibinfo{author}{P.~Dabkowski}, \bibinfo{author}{Y.~Gal},
\newblock \bibinfo{title}{Real time image saliency for black box classifiers},
\newblock in: \bibinfo{booktitle}{Advances in Neural Information Processing
  Systems}, \bibinfo{year}{2017}, pp. \bibinfo{pages}{6967--6976}.
\bibitem[{Zintgraf et~al.(2017)Zintgraf, Cohen, Adel, and Welling}]{22}
\bibinfo{author}{L.~M. Zintgraf}, \bibinfo{author}{T.~S. Cohen},
  \bibinfo{author}{T.~Adel}, \bibinfo{author}{M.~Welling},
\newblock \bibinfo{title}{Visualizing deep neural network decisions: Prediction
  difference analysis},
\newblock \bibinfo{journal}{arXiv preprint arXiv:1702.04595}
  (\bibinfo{year}{2017}).
\bibitem[{Schwab and Karlen(2019)}]{zee12schwab2019cxplain}
\bibinfo{author}{P.~Schwab}, \bibinfo{author}{W.~Karlen},
\newblock \bibinfo{title}{Cxplain: Causal explanations for model interpretation
  under uncertainty},
\newblock \bibinfo{journal}{arXiv preprint arXiv:1910.12336}
  (\bibinfo{year}{2019}).
\bibitem[{Agarwal and Nguyen(2020)}]{zee13agarwal2020explaining}
\bibinfo{author}{C.~Agarwal}, \bibinfo{author}{A.~Nguyen},
\newblock \bibinfo{title}{Explaining image classifiers by removing input
  features using generative models},
\newblock in: \bibinfo{booktitle}{Proceedings of the Asian Conference on
  Computer Vision}, \bibinfo{year}{2020}.
\bibitem[{Cohen et~al.(2021)Cohen, Brooks, En, Zucker, Pareek, Lungren, and
  Chaudhari}]{zee11cohen2021gifsplanation}
\bibinfo{author}{J.~P. Cohen}, \bibinfo{author}{R.~Brooks},
  \bibinfo{author}{S.~En}, \bibinfo{author}{E.~Zucker},
  \bibinfo{author}{A.~Pareek}, \bibinfo{author}{M.~P. Lungren},
  \bibinfo{author}{A.~Chaudhari},
\newblock \bibinfo{title}{Gifsplanation via latent shift: A simple autoencoder
  approach to counterfactual generation for chest x-rays},
\newblock \bibinfo{journal}{arXiv preprint arXiv:2102.09475}
  (\bibinfo{year}{2021}).
\bibitem[{Isola et~al.(2017)Isola, Zhu, Zhou, and Efros}]{27}
\bibinfo{author}{P.~Isola}, \bibinfo{author}{J.-Y. Zhu},
  \bibinfo{author}{T.~Zhou}, \bibinfo{author}{A.~A. Efros},
\newblock \bibinfo{title}{Image-to-image translation with conditional
  adversarial networks},
\newblock in: \bibinfo{booktitle}{Proceedings of the IEEE conference on
  computer vision and pattern recognition}, \bibinfo{year}{2017}, pp.
  \bibinfo{pages}{1125--1134}.
\bibitem[{Zhu et~al.(2017)Zhu, Park, Isola, and Efros}]{16}
\bibinfo{author}{J.-Y. Zhu}, \bibinfo{author}{T.~Park},
  \bibinfo{author}{P.~Isola}, \bibinfo{author}{A.~A. Efros},
\newblock \bibinfo{title}{Unpaired image-to-image translation using
  cycle-consistent adversarial networks},
\newblock in: \bibinfo{booktitle}{Proceedings of the IEEE international
  conference on computer vision}, \bibinfo{year}{2017}, pp.
  \bibinfo{pages}{2223--2232}.
\bibitem[{Shrivastava et~al.(2017)Shrivastava, Pfister, Tuzel, Susskind, Wang,
  and Webb}]{shrivastava2017learning}
\bibinfo{author}{A.~Shrivastava}, \bibinfo{author}{T.~Pfister},
  \bibinfo{author}{O.~Tuzel}, \bibinfo{author}{J.~Susskind},
  \bibinfo{author}{W.~Wang}, \bibinfo{author}{R.~Webb},
\newblock \bibinfo{title}{Learning from simulated and unsupervised images
  through adversarial training},
\newblock in: \bibinfo{booktitle}{Proceedings of the IEEE conference on
  computer vision and pattern recognition}, \bibinfo{year}{2017}, pp.
  \bibinfo{pages}{2107--2116}.
\bibitem[{Kim and Hwang(2016)}]{zee5kim2016deconvolutional}
\bibinfo{author}{H.-E. Kim}, \bibinfo{author}{S.~Hwang},
\newblock \bibinfo{title}{Deconvolutional feature stacking for
  weakly-supervised semantic segmentation},
\newblock \bibinfo{journal}{arXiv preprint arXiv:1602.04984}
  (\bibinfo{year}{2016}).
\bibitem[{Jaeger et~al.(2014)Jaeger, Candemir, Antani, W{\'a}ng, Lu, and
  Thoma}]{zee7jaeger2014two}
\bibinfo{author}{S.~Jaeger}, \bibinfo{author}{S.~Candemir},
  \bibinfo{author}{S.~Antani}, \bibinfo{author}{Y.-X.~J. W{\'a}ng},
  \bibinfo{author}{P.-X. Lu}, \bibinfo{author}{G.~Thoma},
\newblock \bibinfo{title}{Two public chest x-ray datasets for computer-aided
  screening of pulmonary diseases},
\newblock \bibinfo{journal}{Quantitative imaging in medicine and surgery}
  \bibinfo{volume}{4} (\bibinfo{year}{2014}) \bibinfo{pages}{475}.
\bibitem[{zee(????)}]{zee9b}
\bibinfo{title}{Belarus tuberculosis portal},
  \bibinfo{howpublished}{\url{http://tuberculosis.by}}, ????
\bibitem[{Ryoo and Kim(2014)}]{zee8ryoo2014activities}
\bibinfo{author}{S.~Ryoo}, \bibinfo{author}{H.~J. Kim},
\newblock \bibinfo{title}{Activities of the korean institute of tuberculosis},
\newblock \bibinfo{journal}{Osong public health and research perspectives}
  \bibinfo{volume}{5} (\bibinfo{year}{2014}) \bibinfo{pages}{S43--S49}.
\bibitem[{Bakas et~al.(2017)Bakas, Akbari, Sotiras, Bilello, Rozycki, Kirby,
  Freymann, Farahani, and Davatzikos}]{zee6bakas2017advancing}
\bibinfo{author}{S.~Bakas}, \bibinfo{author}{H.~Akbari},
  \bibinfo{author}{A.~Sotiras}, \bibinfo{author}{M.~Bilello},
  \bibinfo{author}{M.~Rozycki}, \bibinfo{author}{J.~S. Kirby},
  \bibinfo{author}{J.~B. Freymann}, \bibinfo{author}{K.~Farahani},
  \bibinfo{author}{C.~Davatzikos},
\newblock \bibinfo{title}{Advancing the cancer genome atlas glioma mri
  collections with expert segmentation labels and radiomic features},
\newblock \bibinfo{journal}{Scientific data} \bibinfo{volume}{4}
  (\bibinfo{year}{2017}) \bibinfo{pages}{1--13}.
\bibitem[{Wang et~al.(2004)Wang, Bovik, Sheikh, and
  Simoncelli}]{zee10wang2004image}
\bibinfo{author}{Z.~Wang}, \bibinfo{author}{A.~C. Bovik},
  \bibinfo{author}{H.~R. Sheikh}, \bibinfo{author}{E.~P. Simoncelli},
\newblock \bibinfo{title}{Image quality assessment: from error visibility to
  structural similarity},
\newblock \bibinfo{journal}{IEEE transactions on image processing}
  \bibinfo{volume}{13} (\bibinfo{year}{2004}) \bibinfo{pages}{600--612}.

\end{thebibliography}
\end{document}